\documentclass[11pt,a4paper]{article}
\usepackage[hyperref]{acl2021}
\usepackage{times}
\usepackage{latexsym}

\usepackage{microtype}

\aclfinalcopy 

\usepackage{prftree, proof, stmaryrd, amssymb}
\usepackage{mathtools, amsthm, float, amstext, cancel}
\usepackage{hyperref, enumerate, enumitem, array, subcaption, multirow, graphicx, graphbox}
\usepackage{color, setspace}
\usepackage{braket}

\theoremstyle{definition}

\title{On the Quantum-like  Contextuality of Ambiguous Phrases} 

\author{ \hspace{15mm} Daphne Wang  ~~ Mehrnoosh Sadrzadeh \\
\hspace{13.1mm}University College London  \\
  \\\And
  \hspace{15mm}\qquad \qquad Samson Abramsky \\
  \hspace{15mm}\qquad \qquad Oxford University \\
  \texttt{}  \\\And
   \hspace{15mm} V\'{i}ctor H. Cervantes \\
   \hspace{15mm}University of Illinois \\ \hspace{13.1mm}at Urbana-Champaign\\
\texttt{}\\}

\date{}

\begin{document}
\maketitle
\begin{abstract}
    Language is contextual as meanings of words are dependent on their contexts. Contextuality is,  concomitantly,  a well-defined concept in quantum mechanics where it is considered a major resource for quantum computations. We investigate whether natural language exhibits any of the quantum mechanics' contextual features. We show that meaning combinations in ambiguous phrases can be modelled in the sheaf-theoretic framework for  quantum contextuality, where they can become possibilistically contextual. Using the framework of Contextuality-by-Default (CbD),  we explore the probabilistic variants of these  and show that CbD-contextuality is also possible. 
\end{abstract}



\section{Introduction}

We start with a peculiar observation: even though  polysemy and homonymy are common phenomena of natural language, i.e. many words have more than one meaning, this does not create a considerable obstacle in our day-to-day comprehension of texts and  conversations. For example, the word \emph{charge} has 40 different senses in English  according to WordNet, however, its meaning in the sentence  \emph{The bull charged.} remains fairly unambiguous. On the other hand, polysemy and word sense disambiguation are computationally difficult tasks and is amongst the challenges faced by linguists \cite{Rayner1986,PickeringFrisson,FRAZIER1990181}.

The emergence of the field of quantum methods in Natural Language Processing  offers  promising leads for introducing quantum methods to  classical NLP tasks, e.g. language modelling \cite{basile-tamburini-2017-towards}, distributional semantics \cite{blacoe-etal-2013-quantum}, mental lexicon \cite{Bruzaetal}, narrative structure \cite{meichanetzidis2020quantum}, emotion detection \cite{Qiuchietal}, and classification  \cite{liu-etal-2013-novel-classifier, Lietal}. 

Distributional semantics is a natural language semantic framework built on the notion of contextuality. Herein,  frequencies of  co-occurrences of words are computed from their contexts and the resulting vector representations  are used in automatic  sense discrimination \cite{schutze}. An issue with this framework is that  the grammatical structure of phrases and sentences is ignored and the focus is mainly on large-scale statistics of data. Oppositely,  even though the interaction between context and syntax has been studied in the past \cite{BarkerShan}, no distributional data has been  considered in them. Finally, distributional and compositional models of language have been proposed \cite{discocat},  small experiments have been implemented on quantum devices \cite{meichanetzidis2020quantum}, and choices of meaning in concept combinations have been analysed using superposition \cite{Conceptual_combinations,piedeleu2015open}. Our work  complements these lines of research by modelling the underlying structure of contextuality using distributional  data.

 We investigate the contextual nature of meaning combinations in ambiguous phrases of natural language, using instances of the data gathered in psycholinguistics \citep{PickeringFrisson,TANENHAUS1979427, Rayner1986},  frequencies  mined from large scale corpora  \cite{BNC,ukWaC},  and models coming from the sheaf-theoretic framework  \cite{AbramskyBrad,logicalBell} and the Contextuality-By-Default (CbD) theory \cite{CbD_ConteNtConteXt}.  We consider phrases  with  two ambiguous words,  in subject-verb and verb-object predicate-argument structures and  find instances of  logical and CbD contextuality.



The structure of the paper is as follows.  We start by introducing the main concepts behind quantum contextuality (section \ref{sec:qc}). We then introduce the sheaf-theoretic framework and logical contextuality (section \ref{sec:background-sheaves}), before applying  it to possibilistic natural language models (section \ref{sec:nlAmbg} and \ref{sec:poss}). In section \ref{sec:prob} and \ref{sec:ns}, we discuss probabilistic models and signalling in natural language respectively. In section \ref{sec:background-CbD}, we offer the possibility of studying contextuality in signalling models via the Contextuality-by-Default framework and discuss two CbD contextual examples that we found. We then close the paper with insights on how to perform a large scale experiment and the possibility of finding more CbD-contextual examples in natural language.

\section{Quantum contextuality}\label{sec:qc}

Early critics of quantum mechanics claimed that quantum theory was not complete \citep{EPR}, but instead was subject to unobserved hidden variables, and claimed that any physical theory should satisfy local realism. By local realism, one means that in a ``complete'' physical theory, the global behaviour of a system is entirely, and deterministically, 
fixed by a set of local variables. However, the well-known Bell theorem \citep{bell}, supported by experimental data \citep{loopholefreeBell}, shows that a description of quantum mechanics cannot comply with local realism; if quantum systems need to have a ``reality'' independent of the observers (realism), one should allow interactions between systems to be unrestricted spatially (non-local).

The Bell inequality offers a proof by contradiction that one cannot extend the probabilistic models obtained from observations of quantum systems to a deterministic hidden-variable model. In \citet{KS}, the authors prove a stronger statement about the existence of hidden-variable models via a logical argument. This more general result provides a description of \emph{contextuality} as it is understood in quantum mechanics.


\section{Presheaves and logical contextuality}\label{sec:background-sheaves}


The sheaf-theoretic framework of contextuality starts from the observation that contextuality in quantum mechanics translates to the impossibility of finding a global section in special presheaves. In other words, a model is contextual if some of its \emph{local} features cannot be extended \emph{globally}.
\begin{figure}[t!]
    \centering
    \begin{subfigure}[b]{\linewidth}
        \centering
        \begin{tabular}{cc|cccc}
            $A$ & $B$ & (0,0) & (0,1) & (1,0) & (1,1) \\\hline
            $a$ & $b$ & $1/2$ & $0$ & $0$ & $1/2$\\
            $a$ & $b'$ & $3/8$ & $1/8$ & $1/8$ & $3/8$\\
            $a'$ & $b$ & $3/8$ & $1/8$ & $1/8$ & $3/8$\\
            $a'$ & $b'$ & $1/8$ & $3/8$ & $3/8$ & $1/8$\\
        \end{tabular}
        \caption{Probability distributions}
    \end{subfigure}
    \begin{subfigure}[b]{\linewidth}
        \centering
        \includegraphics[width=.75\linewidth]{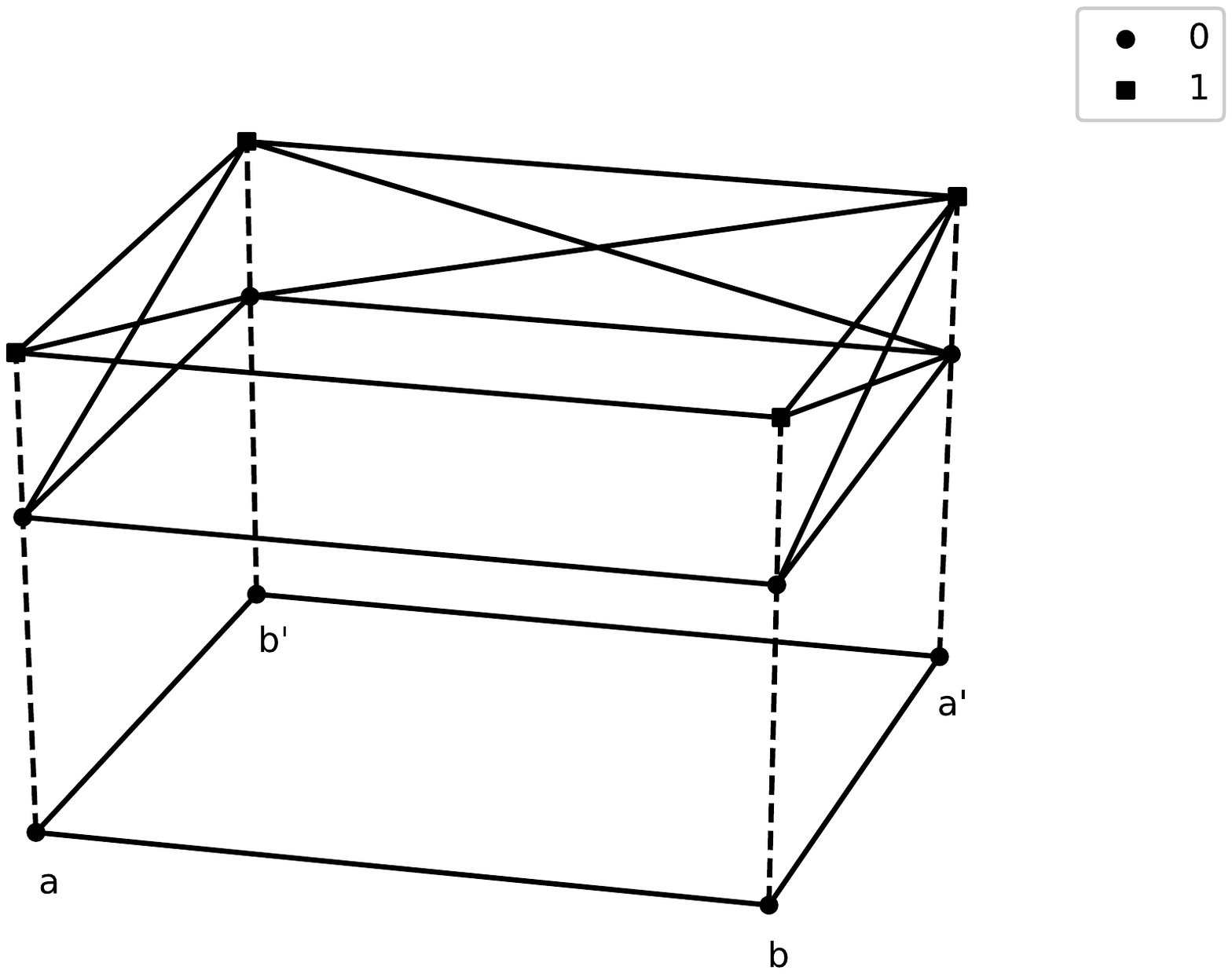}
        \caption{Bundle diagram of the logical model\label{subfig:BellBundle}}
    \end{subfigure}
    \caption{Empirical model associated with the measurement of the bipartite state $\ket{\Psi} = \left.\frac{1}{\sqrt{2}}\middle(\ket{00}+\ket{11}\right)$ with local measurements $a,b = \ket{1}\bra{1}_{A,B}$ and $a',b' = \ket{\phi}\bra{\phi}_{A,B}$ where $\ket{\phi} = \frac{\sqrt{3}}{2}\ket{0} + i\frac{1}{2}\ket{1}$.\label{fig:exampleBell}}
\end{figure}

\paragraph{}The presheaves considered in the framework developed by Abramsky et. al.~\cite{AbramskyBrad,bundle} are so-called distribution presheaves on events. An \emph{empirical model} corresponds to the experiment that is undertaken; it consists of the list of measurements that can be made, which measurements can be made together and what are the associated probability distributions. For example, Fig.~\ref{fig:exampleBell} can represent a standard Bell experiment where the list of measurements is the list of all local measurements that can be made on the two qubits involved in the experiment, i.e. $\{a,a',b,b'\}$, under  the condition that each laboratory performs exactly one measurement at each run of the experiment, e.g. $(a,b)$ can be a joint measurement, but $(a,a')$ cannot. The distribution presheaf then associates the observed (or theoretical) probabilities for the global measurement outcomes, that is, the joint outcomes of both parties in a Bell scenario, for each measurement context. In this framework, a \emph{global assignment} corresponds to an assignment of an outcome for every local measurement. A \emph{global section} will on the other hand represent a distribution defined on all global assignments, which is consistent with all the observed probabilities.

The framework of Abramsky et al.~\cite{AbramskyBrad,bundle, logicalBell} also introduces a stronger type of contextuality, called \emph{logical} (or \emph{possibilistic}) \emph{contextuality}. Indeed, they have found that the contextuality of some systems can be established from the \emph{support} of each of the context-dependent distributions. These are referred to as \emph{possibilistic empirical models}. In these models, we are only interested in whether an outcome of a local measurement (given a global measurement context) is \emph{possible}.
A consistent global assignment will then be an assignment of a possible outcome to every measurement, and hence can be represented by a logical statement about a subsystem.  A global section will then be a disjunction of consistent global assignments that describes the entirety of the model. Hence, one can prove logical contextuality, i.e. the impossibility of being able to write such a logical statement about the system, by finding a locally possible outcome that cannot be extended to a consistent global assignment.

For small systems, it is convenient to represent possibilistic models by bundle diagrams \citep{bundle}. In these diagrams, we represent each of the local measurements by a vertex. There is an edge\footnote{More generally simplices if multiple measurements are carried out simultaneously.} between every two of these vertices if the joint measurement is possible. We then depict, for each individual measurements, the set of possible outcomes as a set ``sitting'' on top of the associated vertex. Similarly,  an edge is added between two of the ``outcome''-vertices if the joint measurement has a non-zero probability (e.g. see Fig.~\ref{subfig:BellBundle}). 
In particular, global assignments can be seen in these bundle diagrams as shapes going through exactly one outcome for each of the measurements that mirror the structure of the base (measurements). In Fig.~\ref{subfig:BellBundle} for example, 
global assignments correspond to connected loops.

The sheaf-theoretic framework relies on the fact that the described distribution presheaf is indeed a presheaf. That is,  the distributions associated with measurement contexts that intersect at a local measurement (i.e. two contexts where at least one party performs the same measurement) agree on their restrictions. These are here defined as the marginals of the distributions of interest. This requirement coincides with the \emph{non-signalling} condition in quantum mechanics. 
This condition is stated for possibilistic models by requiring that the supports of intersecting distributions coincides.

As we will see, many empirical models from natural language will be \emph{signalling}.
That is also the case for many behavioural and psychological experiments (see e.g. \citet{Conceptual_combinations,CbD_Behavioural_Social_systems}), and in fact, there is no reason why natural language systems \emph{should} be non-signalling and we will discuss this issue in sections \ref{sec:ns}.

\section{Contextuality and ambiguity in natural language}\label{sec:nlAmbg}

\begin{figure*}[t!]
    \centering
    \begin{subfigure}[b]{.3\linewidth}
        \centering
        \includegraphics[width=\linewidth]{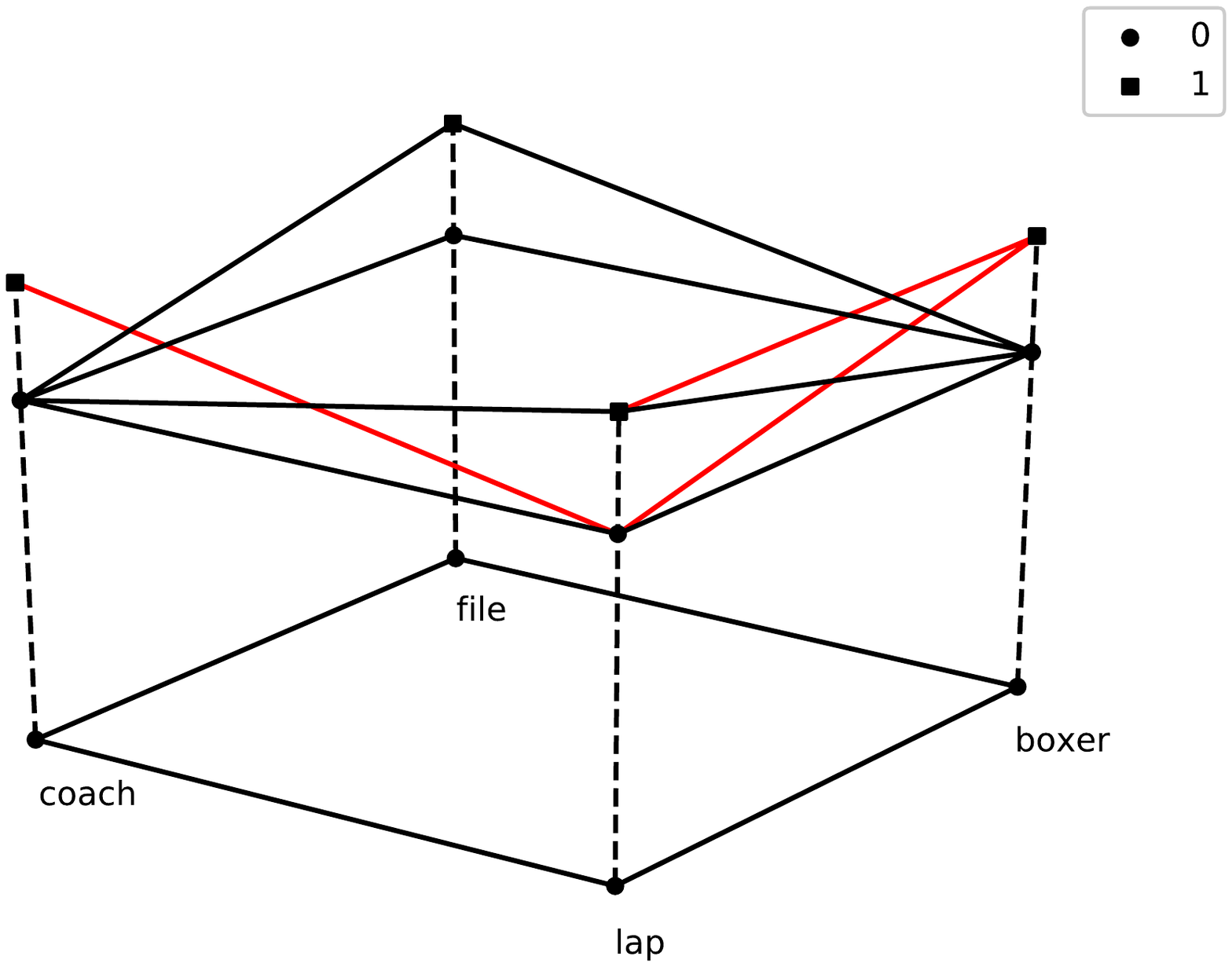}
        \caption{$\{$\emph{coach}, \emph{boxer}$\}\times\{$\emph{lap}, \emph{file}$\}$\label{subfig:coachlapboxerfile}}
    \end{subfigure}
    \begin{subfigure}[b]{.3\linewidth}
        \centering
        \includegraphics[width=\linewidth]{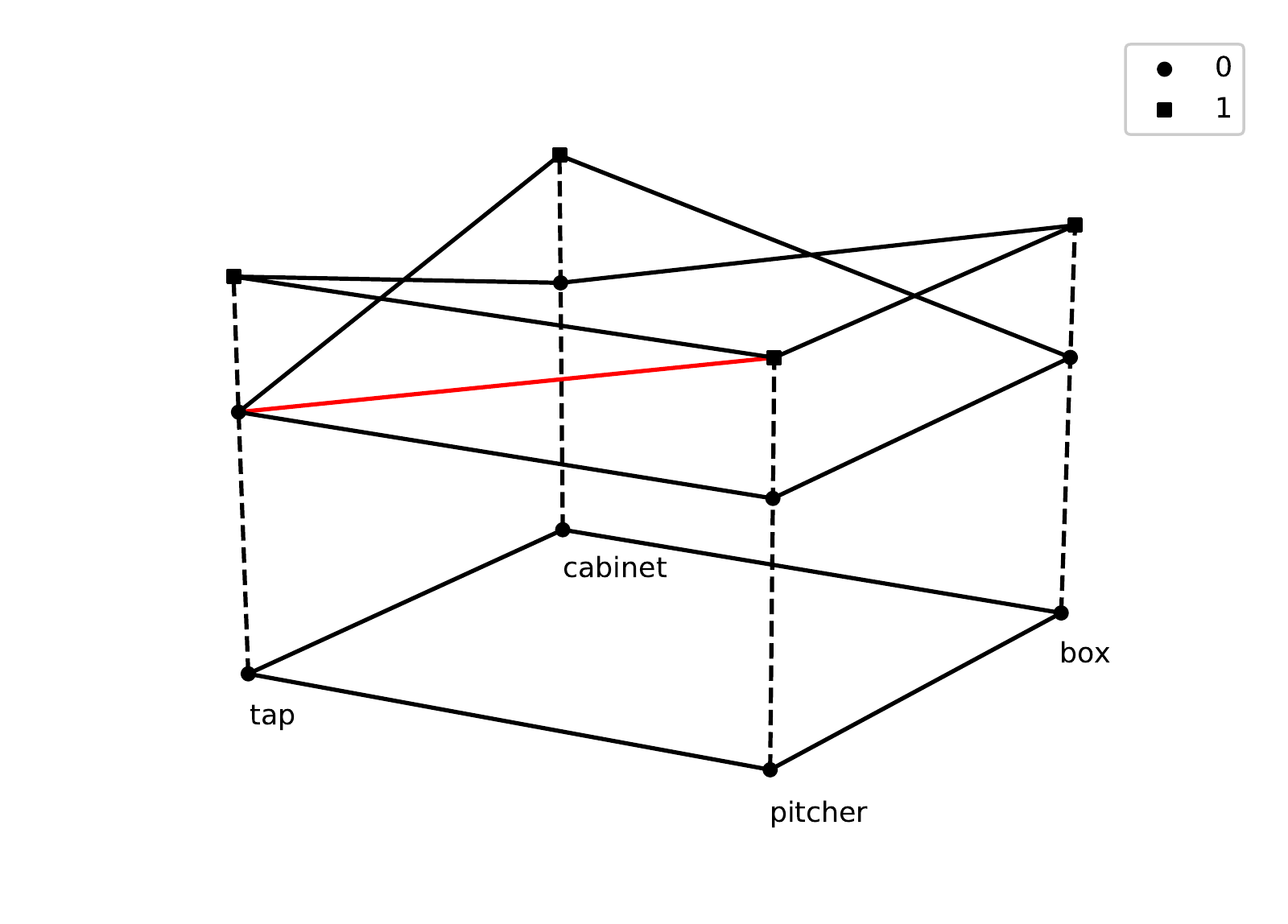}
        \caption{$\{$\emph{tap}, \emph{box}$\}\times\{$\emph{pitcher}, \emph{cabinet}$\}$\label{subfig:tappitcherboxcabinet}}
    \end{subfigure}%
    \begin{subfigure}[b]{.3\linewidth}
        \centering
        \includegraphics[width=\linewidth]{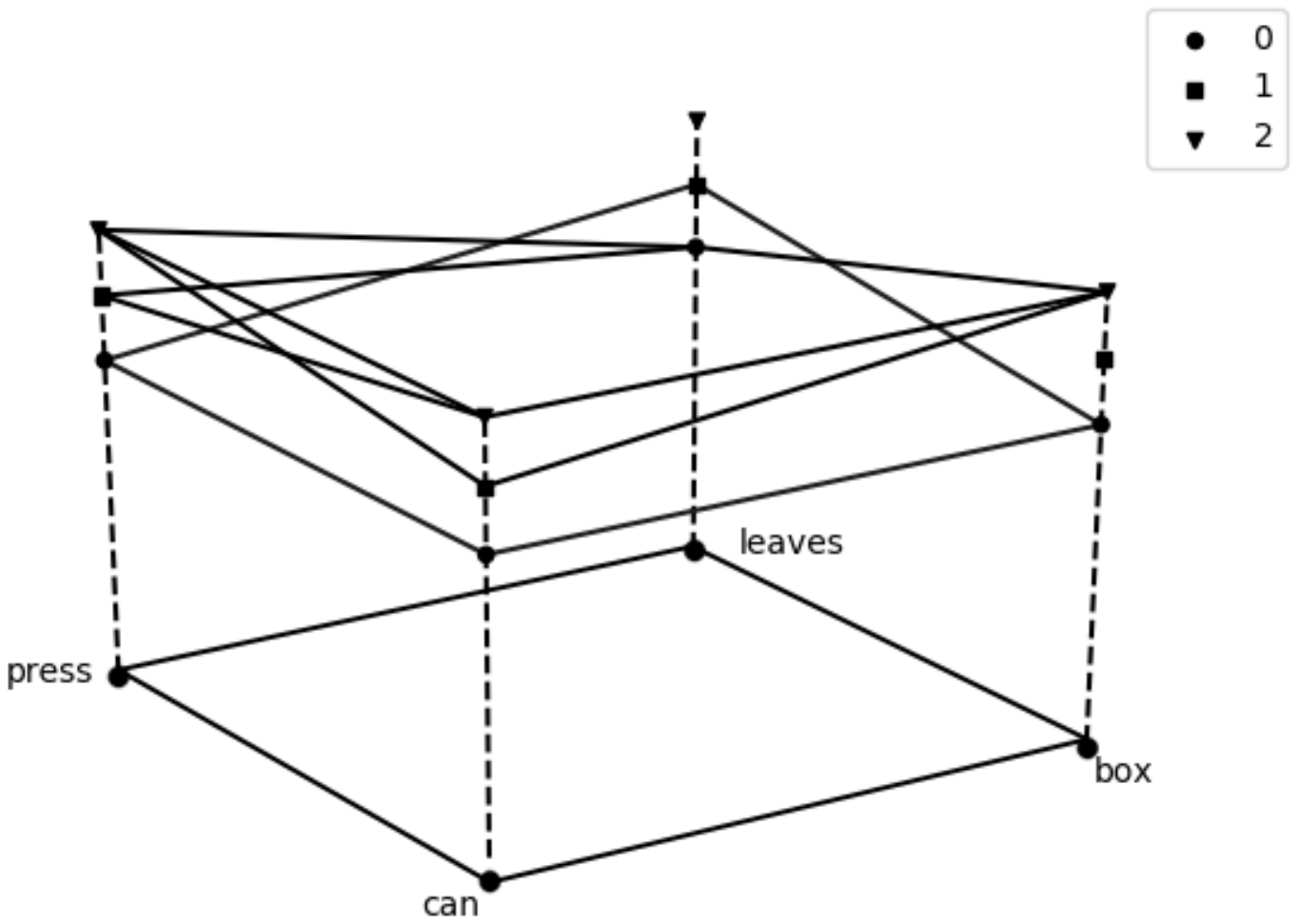}
        \caption{$\{$\emph{press}, \emph{box}$\}\times\{$\emph{can}, \emph{leaves}$\}$\label{subfig:presscanboxleaves}}
\end{subfigure}
    \caption{Instances of bundle diagrams arising from ambiguous phrases. The local assignments which cannot be extended to a global one are depicted in red.\label{fig:bundles}}
\end{figure*}

We are interested in studying the influence of the context on the process of meaning selection in ambiguous phrases. Indeed, homonymy and polysemy in natural language give rise to an interpretation for context-dependent probability distributions. Probabilities will correspond to the likelihood that a certain meaning of a word is selected in the context of interest. By analogy with quantum contextuality,  existence of contextual natural language examples  confirms that the context in which words are found plays a non-trivial role in the selection of an appropriate interpretation for them and the following question  arises: given that  a certain interpretation of a word is selected within a certain context, can we use this information to deduce how the same word may be interpreted in a different context (e.g.  in different phrases) in the  corpus?

Our intuition is that this is not the case. Consider the ambiguous adjective \emph{green}: this either refers to the colour of its modifier (e.g. \emph{a green door}), or the environmental-friendly nature of it (e.g. \emph{the Green party}). Now, if we consider an unambiguous adjective such as \emph{new}, then trivially, the same interpretations of \emph{new} can be selected in both of the phrases \emph{new paint} and \emph{new policy}.  This, however,  does not imply that the same interpretations of \emph{green} will be selected in \emph{green paint} and \emph{green policy}. With this intuition in mind, we start by considering the basic structure of ambiguous phrases of  English by considering only the support of probability distributions attributed to these phrases and for now appeal to our common sense to determine the values of these supports.

\begin{figure}[t!]
    \centering
    \includegraphics[height=3cm]{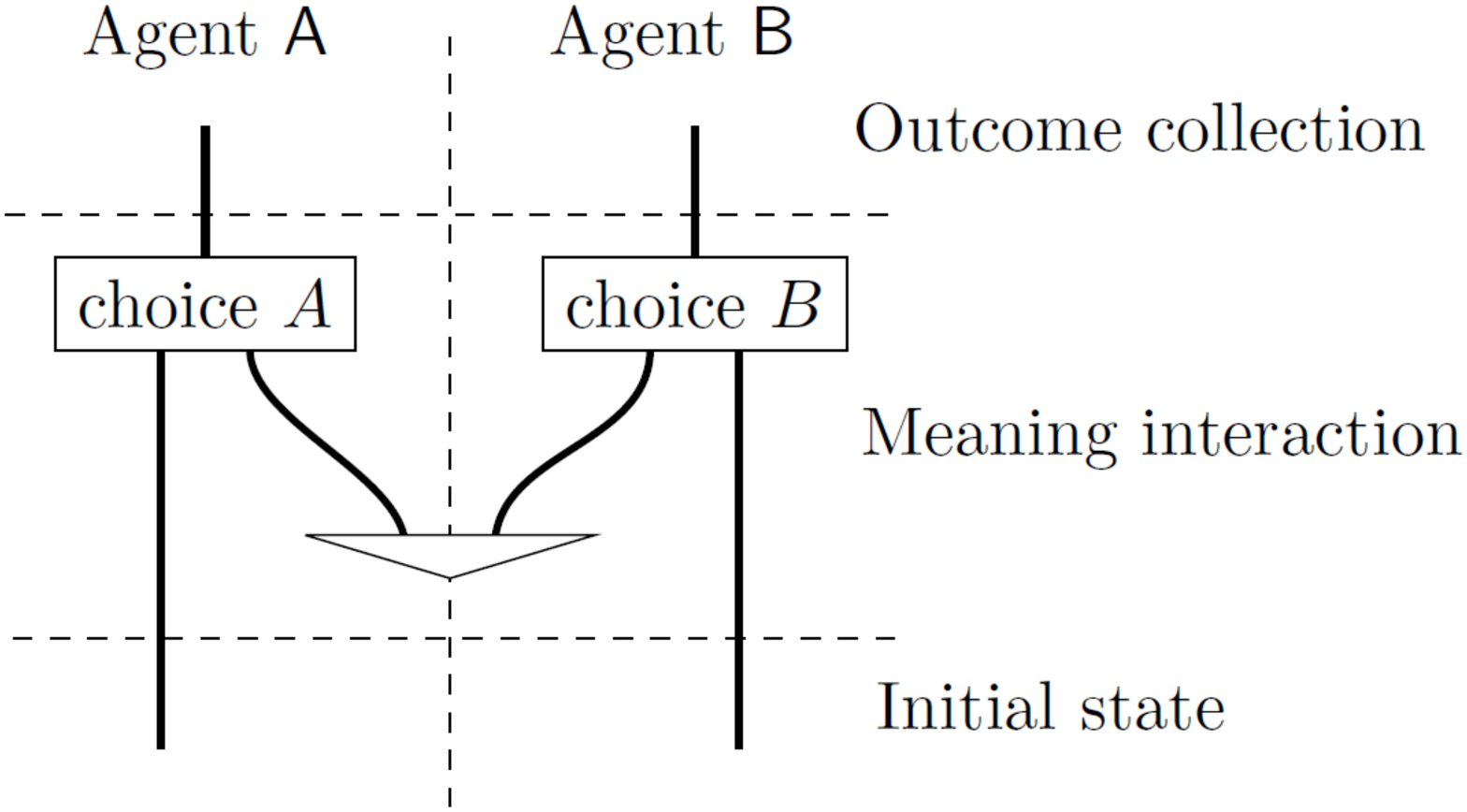}
    \caption{Example of a 2-words scenario. The state (triangle) represents the predefined conditions of the interaction (e.g. $verb-object$).\label{fig:BellSC}}
\end{figure}

In the first part of the paper, we consider a structure  similar to Bell scenarios with multiple parties, or agents, each of which will choose one measurement context from a predetermined set. A ``measurement'' will be associated with each word and will return the activated meaning according to a fixed encoding. For example the two meanings of \emph{green} could be encoded as: \emph{relative to colour} $\mapsto 0$, \emph{environmental-friendly} $\mapsto 1$. In a given context, multiple ambiguous words will be allowed to ``interact'' and form a phrase. A measurement context will then be labelled by the words in this phrase. 
The interaction will  be dictated by some predetermined rules, such as which part-of-speech each word will correspond to. For each global measurement context, the recorded activated meanings will then form a joint distribution. These distributions can  be represented in the form of an empirical model as described in section \ref{sec:background-sheaves}. In order to obtain a valid empirical model, all the possible combinations of words need to make  sense. For example, take two parties $\mathsf{A}$ and $\mathsf{B}$ such that $\mathsf{A}$ chooses an adjective in the set $\{$\emph{green},\emph{new}$\}$ and $\mathsf{B}$ chooses its modifier within the set $\{$\emph{paint}, \emph{policy}$\}$.  All the combinations of \(\mathsf{A}\) and \(\mathsf{B}\) are possible, i.e.  phrases \emph{green paint}, \emph{green policy}, \emph{new paint} and \emph{new policy} all make sense and can indeed be found in natural language corpora. However, if the set of adjectives is changed to $\{$\emph{blue}, \emph{new}$\}$, we will face a problem since the phrase \emph{blue policy} does not make much sense and we could not find any occurrence of it in the corpora considered in this paper.\footnote{One may imagine a metaphorical meaning of this phrase, e.g. when referring to a  \emph{depressing policy.} In this paper, however, we  work with non metaphorical  meanings in order to keep the hand annotations of interpretations manageable.} In order to keep the models and computations simple, we work with 2-word phrases, where each word of the phrase is ambiguous. From the analogy with Bell scenarios, this means that we are working with bipartite scenarios (see Fig. \ref{fig:BellSC}). The set of ambiguous words is taken from experimental data sets from the studies: \citet{PickeringFrisson,Rayner1986,TANENHAUS1979427}. 



In sections \ref{subsec:poss_rank2} and \ref{subsec:prob_rank2}, we introduce another kind of experiment which departs from Bell scenarios. Measurements of these examples have the same interpretation as before, but the focus is on combinations involving a single verb and a single noun for which both of subject-verb and verb-object phrases are possible. 
This structure is analogous to the scenario in behavioural sciences for the ``Question Order effect''
\citep{Wang.2013.Quantum}. In the sheaf-theoretic framework measurement contexts are dictated only by the choices of local measurements and we face two possibilities when modelling these examples. In the first possibility, one can  consider the two contexts subject-verb/verb-object as disjoint and as a result lose some semantic information. 
This is because,  for example, \emph{adopt} in \emph{adopt boxer} would be treated as completely unrelated to \emph{adopt} in \emph{boxer adopts}. In this case, all such systems will be trivially non-contextual, as there will be no intersecting local measurements. In our paper, on the other hand, we choose a second possibility and decide to keep the semantic information  but as a result any system for which the distribution arising from the verb-object context differs from the one associated with subject-verb context will be signalling. This type of model  does not easily lend itself to a sheaf-theoretic analysis but admits a straightforward  CbD analysis\footnote{We are not aware of any theoretical reason why Bell-scenario-like models could not be CbD-contextual, none, however, have been found using the corpus mining methodology of  this paper.}.

\section{Possibilistic  examples}\label{sec:poss}

We demonstrate the methodology by choosing  three sets of phrases from the sets  considered by \citet{DaphneWangMRes} as well as two verb-object/subject-verb examples. For each of these phrases, we tabulate how we encoded the  meanings of each word, provide an empirical Bell-style table for the possibilistic cases and outline the different types of contextual features each example demonstrates. 

\subsection{$\{$\emph{coach}, \emph{boxer}$\}\times\{$\emph{lap}, \emph{file}$\}$}\label{subsec:poss_coachlapboxerfile}
\begin{figure}[ht]
    \centering
    \begin{subfigure}[b]{\linewidth}
        \centering
        \scalebox{.8}{
            \begin{tabular}{|c||m{.1\linewidth} m{.1\linewidth} m{.1\linewidth} m{.2\linewidth} |}
            \hline\multirow{2}{*}{Encoding}  & \multicolumn{4}{c|}{Meanings of}\\
              &\emph{coach} & \emph{boxer} & \emph{lap} & \emph{file}\\\hline
             0 &\emph{sport} & \emph{boxing} & \emph{run} & \emph{document}\\
             1 &\emph{bus} & \emph{dog} & \emph{drink} & \emph{smooth} \\\hline
        \end{tabular}}
        \caption{Encoding of meanings of \emph{coach}, \emph{boxer}, \emph{lap} and \emph{file}.}\label{subfig:encoding_coachlapboxerfile}
        \end{subfigure}
        \begin{subfigure}[b]{\linewidth}
        \small
            \centering
            \begin{tabular}{cc|cccc}
                 $subject$ & $verb$ & (0,0) & (0,1) & (1,0) & (1,1) \\ \hline
                 $coach$ & $lap$ & 1 & 1 & 1 & 0\\
                 $coach$ & $file$ & 1 & 1 & 0 & 0\\
                 $boxer$ & $lap$ & 1 & 1 & 1 & 1\\
                 $boxer$ & $file$ & 1 & 1 & 0 & 0
            \end{tabular}
            \caption{Empirical model\label{subfig:posstable_coachlapboxerfile}}
            \end{subfigure}
    \caption{Possibilistic model associated with the subject-verb model $\{$\emph{coach}, \emph{boxer}$\}\times\{$\emph{lap}, \emph{file}$\}$.\label{fig:poss_coachlapboxerfile}}
\end{figure}
We start with two subject-verb phrases where both of the subjects and  both of the verbs are ambiguous. The verbs are \emph{lap} and \emph{file}, which can be understood as drinking a liquid (e.g. \emph{the dog lapped the water}) or going past someone on a  track (e.g. \emph{the runner lapped their competitor}) for \emph{lap}, and storing information (e.g. \emph{filing a complaint}) or smoothing surfaces with a tool (e.g. \emph{filing nails or teeth}) for \emph{file}. The nouns \emph{coach} and \emph{boxer} mean a person who trains athletes (e.g. \emph{a sport coach}) or a type of  bus (e.g. \emph{a coach trip}), and a person practising boxing (e.g. \emph{a heavyweight boxer}), or a specific dog breed respectively. This example is modelled possibilistically  in  Fig. \ref{subfig:posstable_coachlapboxerfile} and depicted in  the bundle diagram of Fig. \ref{subfig:coachlapboxerfile}. Not all of the local assignments can be extended to a global one, for example, the assignment $coach \mapsto bus$ is possible in the phrase \emph{the coach laps}, but this assignment cannot be extended in the phrase \emph{the coach files}.

This apparent ``contextuality'', however, is entirely due to the fact that the model is possibilistically signalling and  can be seen by the fact that the support of the contexts \emph{the coach lap} and \emph{the coach files}, restricted to the measurement \emph{coach} do not coincide ([\emph{coach} $\mapsto$ \emph{bus}] $\in \left.coach~lap\right|_{coach}$ but [\emph{coach} $\mapsto$ \emph{bus}] $\not\in \left.coach~file\right|_{coach}$). Hence, we cannot judge the contextuality of this model in the sheaf-theoretic framework.

\subsection{$\{$\emph{tap}, \emph{box}$\}\times\{$\emph{pitcher}, \emph{cabinet}$\}$}\label{subsec:poss_tappitcherboxcabinet}
We now consider an empirical model which is possibilistically non-signalling, and in fact contextual. 
%
This model deals with  two verb-object phrases where the verbs are  $\{$\emph{tap}, \emph{box}$\}$, and their objects are $\{$\emph{pitcher}, \emph{cabinet}$\}$. Here, \emph{tap} is taken to mean either gently touching (e.g. \emph{tapping somebody on the shoulder}) or secretly recording (e.g. \emph{tapping phones}); other meanings of the verb \emph{tap} exist (e.g. doing tap dancing, tapping resources, etc.), but since these other meanings are irrelevant in the phrases of interest, we restrict ourselves to these two  meanings. In addition, the verb \emph{box} is understood as putting in a container and practising boxing. Again, other meanings of the verb \emph{to box} exist,  but as before,  we worked with  two dominant meanings and ignored the rest. The noun \emph{cabinet} either represents a governmental body (e.g. \emph{the Shadow Cabinet}) or a piece of furniture, and finally the noun \emph{pitcher} either refers to a jug or a baseball player. As we can see in Fig. \ref{subfig:tappitcherboxcabinet}, the assignment \emph{tap} $\mapsto$ \emph{touch} cannot be extended to a global assignment and is therefore possibilistically contextual.

\begin{figure}[ht]
    \begin{subfigure}[b]{\linewidth}
        \centering
        \scalebox{.8}{
            \begin{tabular}{|c||p{.1\linewidth} p{.25\linewidth} p{.22\linewidth} p{.22\linewidth} |}
            \hline\multirow{2}{*}{Encoding}  & \multicolumn{4}{c|}{Meanings of}\\
            &\emph{tap} & \emph{box} & \emph{cabinet} & \emph{pitcher}\\\hline
            0 &\emph{touch} & \emph{put in boxes} &\emph{government} & \emph{jug}\\
            1 &\emph{record} & \emph{fight} & \emph{furniture} & \emph{baseball player}\\\hline
        \end{tabular}}
    \caption{Encoding of meanings of \emph{tap}, \emph{box}, \emph{cabinet} and \emph{pitcher}.\label{subfig:encoding_tappitcherboxcabinet}}
    \end{subfigure}
    \begin{subfigure}[b]{\linewidth}
    \small
        \centering
        \begin{tabular}{cc|cccc}
             $verb$ & $object$ & (0,0) & (0,1) & (1,0) & (1,1) \\ \hline
             $tap$ & $pitcher$ & 1 & 1 & 0 & 1\\
             $tap$ & $cabinet$ & 0 & 1 & 1 & 0\\
             $box$ & $pitcher$ & 1 & 0 & 0 & 1\\
             $box$ & $cabinet$ & 0 & 1 & 1 & 0
        \end{tabular}
        \caption{Empirical model}
        \end{subfigure}
        \caption{Possibilistic model associated with the verb-object model $\{$\emph{tap}, \emph{box}$\}\times\{$\emph{pitcher}, \emph{cabinet}$\}$.}
\end{figure}

As we move to section \ref{sec:prob} and   mine probability distributions from corpus for this same model, we see that this possibilistically non-signalling model becomes probabilistically signalling.

\subsection{$\{$\emph{press}, \emph{box}$\}\times\{$\emph{can}, \emph{leaves}$\}$}\label{subsec:poss_presscanboxleaves}
In this model, each word has multiple grammatical types and different meanings  as follows:
\begin{itemize}[label=-, nosep]
    \item \emph{to press} (v): Exert pressure upon something
    \item \emph{press} (n): Media which publishes newspapers and magazines
    \item \emph{press} (n): Device used to apply pressure. (e.g. \emph{They used to use printing presses before the invention of printers.})
    \item \emph{to box} (v): To put in a box
    \item \emph{to box} (v): To fight, to practice boxing
    \item \emph{box} (n): Container
    \item \emph{can} (n) : Tin container
    \item \emph{to can} (v): To preserve food in a can (e.g. \emph{He cans his own sardines.})
    \item \emph{can} (auxiliary): To be able to
    \item \emph{leaves} (v) : Conjugated form of  \emph{to leave}
    \item \emph{leaves} (n): Plural of \emph{leaf}
\end{itemize}
As we can see in the bundle diagram associated with the model (Fig. \ref{subfig:presscanboxleaves}), the marginals of the possibilistic distributions which share a local measurement have the same support,  making this model possibilistically non-signalling. In addition, every local section can be extended to a global assignment, which makes the model non-contextual. In section \ref{subsec:prob_presscanboxleaves} and section \ref{sec:background-CbD}, we endeavour to see whether this model is probabilistically contextual.

\begin{figure}[ht]
    \centering
    \begin{subfigure}[b]{\linewidth}
        \centering
        \scalebox{.8}{
        \begin{tabular}{|c||m{.15\linewidth} p{.25\linewidth} m{.15\linewidth} m{.13\linewidth} |}
            \hline\multirow{2}{*}{Encoding}  & \multicolumn{4}{c|}{Meanings of}\\
          & \emph{press} & \emph{box} & \emph{can} & \emph{leaves}\\\hline
         0 & \emph{push} & \emph{put in boxes} & \emph{tin} & \emph{leave}\\
         1 & \emph{media} & \emph{fight} & \emph{preserve} & \emph{leaf}\\
         2 & \emph{machine} & \emph{container} & \emph{able to} & $\star$\\\hline
        \end{tabular}}
    \caption{Encoding of meanings of \emph{press}, \emph{box}, \emph{can} and \emph{leaves}.\label{subfig:encoding_presscanboxleaves}}
    \end{subfigure}
    \begin{subfigure}[b]{\linewidth}
        \centering
        \scalebox{.65}{
    \hspace{-1.5cm}    \begin{tabular}{cc|ccccccccc}
            $\mathsf{A}$ & $ \mathsf{B}$ & (0,0) & (0,1) & (0,2) & (1,0) & (1,1) & (1,2) & (2,0) & (2,1) & (2,2) \\\hline
            $press$ & $can$ & $1$ & $0$ & $0$ & $0$ & $0$ & $1$ & $0$ & $1$ & $1$\\
            $press$ & $leaves$ & $0$ & $1$ & $0$ & $1$ & $0$ & $0$ & $1$ & $0$ & $0$\\
            $box$ & $can$ & $1$ & $0$ & $0$ & $0$ & $0$ & $0$ & $0$ & $1$ & $1$\\
            $box$ & $leaves$ & $0$ & $1$ & $0$ & $0$ & $0$ & $0$ & $1$ & $0$ & $0$\\
        \end{tabular}}
        \caption{Empirical model}
    \end{subfigure}
    \caption{Possibilistic model associated with the model $\{$\emph{press}, \emph{box}$\}\times\{$\emph{can}, \emph{leaves}$\}$.}
\end{figure}

\subsection{Subject-verb v. Verb-object}\label{subsec:poss_rank2}
\begin{figure}[ht]
    \centering
    \begin{subfigure}[b]{.4\linewidth}
        \centering
        \includegraphics[width=\linewidth]{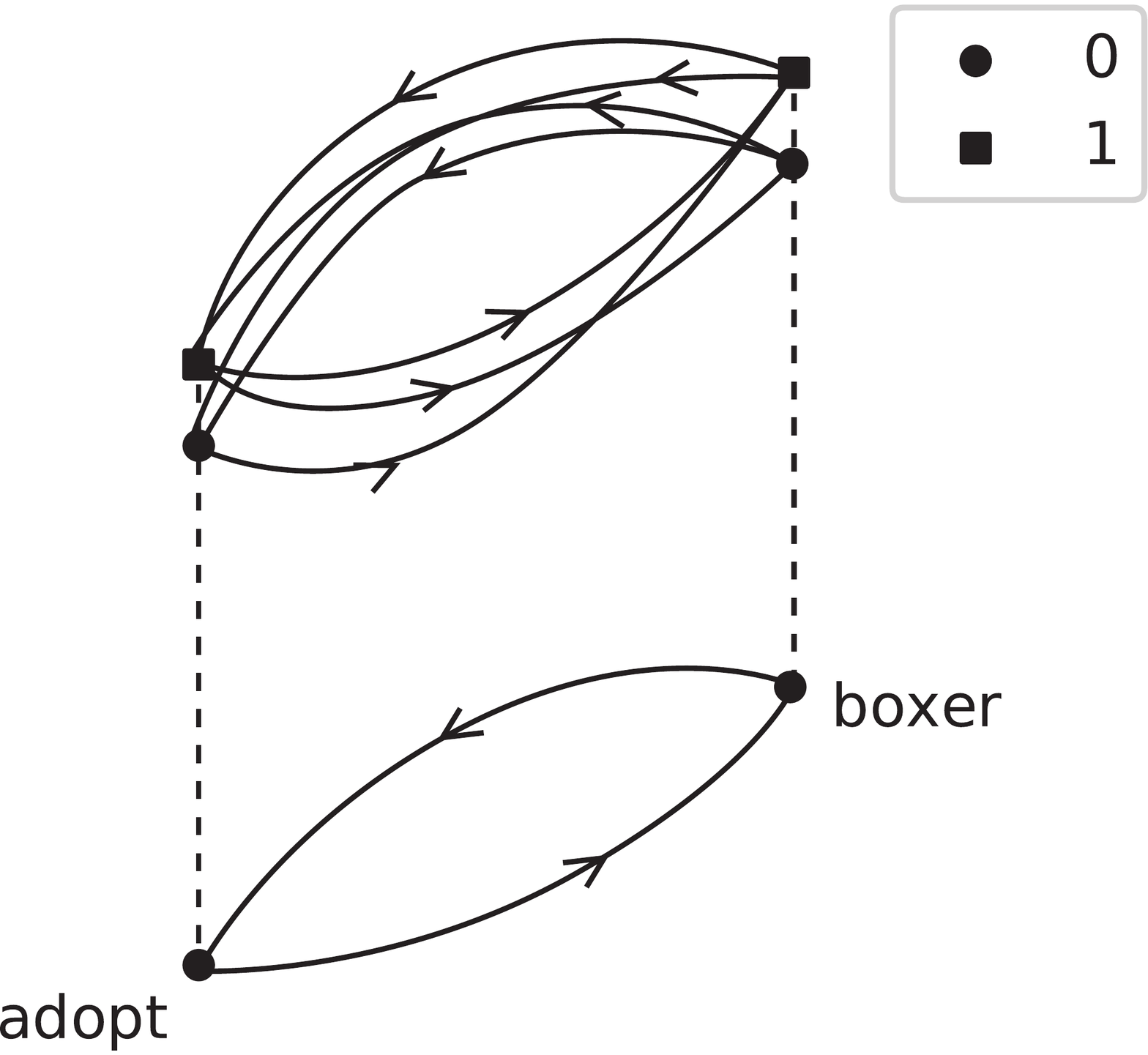}
        \caption{\emph{adopt boxer/boxer adopts}}
    \end{subfigure}\quad%
    \begin{subfigure}[b]{.4\linewidth}
        \centering
        \includegraphics[width=\linewidth]{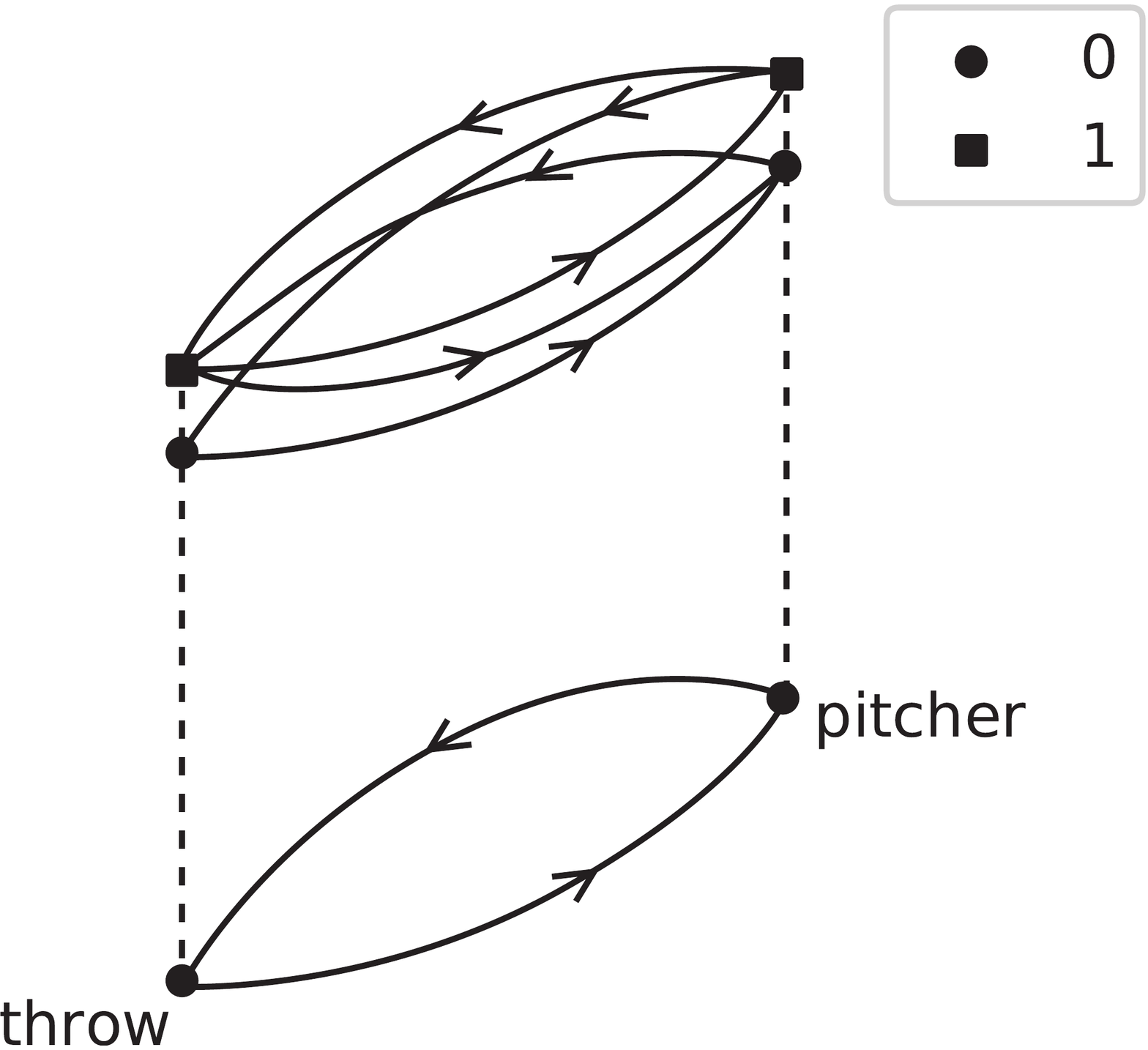}
        \caption{\emph{throw pitcher/pitcher throws}}
    \end{subfigure}
    \caption{Bundle diagrams of the two noun-verb pairs with contexts verb-object and subject-verb. The encoding of the nouns are the same as in Figs.~\ref{subfig:encoding_coachlapboxerfile} and \ref{subfig:encoding_tappitcherboxcabinet}; for verbs, outcomes 0 and 1 represent literal and figurative meanings, respectively. The  measurement contexts (verb-object or subject-verb) are depicted by arrows on the associated edges.\label{fig:bundle_rank2}}
\end{figure}

We now introduce two models for which both of subject-verb and verb-object contexts are possible and provide two  examples.  These are the combinations \emph{adopt boxer}/\emph{boxer adopts} and \emph{throw pitcher}/\emph{pitcher throws}, where \emph{boxer} and \emph{pitcher} are defined as in sections \ref{subsec:poss_coachlapboxerfile} and \ref{subsec:poss_tappitcherboxcabinet} respectively, and the verbs \emph{adopt} and \emph{throw} can either take  literal (e.g. \emph{adopt a child or a pet}, \emph{throwing a projectile}) or figurative (e.g. \emph{adopt a new feature}, or \emph{throwing shadows}) interpretations. The possibilistic models associated with these examples are depicted in Fig. \ref{fig:poss_rank2} and in the bundle diagrams of  Fig. \ref{fig:bundle_rank2}. The models are signalling  and hence, a sheaf-theoretic analysis would not be possible.

\begin{figure}[ht]
    \centering
    \begin{subfigure}[b]{\linewidth}
        \small
        \centering
        \begin{tabular}{c|cccc}
             (\emph{adopt}, \emph{boxer}) & (0,0) & (0,1) & (1,0) & (1,1) \\ \hline
             \emph{adopt} $\rightarrow$ \emph{boxer} & 0 & $1$ & $1$ & $1$\\
             \emph{adopt} $\leftarrow$ \emph{boxer} & $1$ & $1$ & $1$ & $1$
        \end{tabular}
        \caption{\emph{adopt boxer/boxer adopts}}\label{subfig:poss_adoptboxer}
    \end{subfigure}
    \begin{subfigure}[b]{\linewidth}
        \small
        \centering
        \begin{tabular}{c|cccc}
             (\emph{throw}, \emph{pitcher}) & (0,0) & (0,1) & (1,0) & (1,1) \\ \hline
             \emph{throw} $\rightarrow$ \emph{pitcher} & $1$ & $0$ & $1$ & $1$\\
             \emph{throw} $\rightarrow$ \emph{pitcher} & 0 & $1$ & $1$ & $1$
        \end{tabular}
        \caption{\emph{throw pitcher/pitcher throws}}\label{subfig:poss_throwpitcher}
    \end{subfigure}
    \caption{Empirical models for the pairs of words examples. Here, the different contexts are depicted as follows: \emph{verb}$\rightarrow$ \emph{noun} corresponds to the verb-object context while \emph{verb}$\leftarrow$\emph{noun} corresponds to the subject-verb phrase. The outcomes labels are the same for  both contexts; for example $(0,1)$ in (a) means \emph{adopt}$\mapsto 0$, \emph{boxer}$\mapsto 1$ for both contexts.\label{fig:poss_rank2}}
    
\end{figure}


\section{Probabilistic variants} \label{sec:prob}

We consider the same examples as in the previous section, but  from a probabilistic point of view.  The probability distributions are obtained from the British National Corpus \citep{BNC} and UKWaC \citep{ukWaC}.   BNC is an open-source text corpus comprising of 100 million words, spread across documents of different nature (including press articles, fiction, transcription of spoken language, and academic publications). UKWaC  is a 2 billion word corpus constructed from the Web limiting the crawl to the .uk domain.  Both BNC and UKWaC are part-of-speech tagged, hence, they provide grammatical relations and the lemma forms of words. The semantic interpretation of the words and phrases are absent from these corpora and had to be decided by the authors manually.

\subsection{$\{$\emph{coach}, \emph{boxer}$\}\times\{$\emph{lap}, \emph{file}$\}$}\label{subsec:prob_coachlapboxerfile}
\begin{figure}[ht]
    \centering
    \small
    \begin{tabular}{cc|cccc}
        $subject$ & $verb$ & (0,0) & (0,1) & (1,0) & (1,1)\\\hline
        $coach$ & $lap$ & $2/11$ & $7/11$ & $2/11$ & $0$\\
        $coach$ & $file$ & $43/44$ & $1/44$ & $0$ & $0$\\
        $boxer$ & $lap$ & $11/53$ & $22/53$ & $8/53$ & $12/53$ \\
        $boxer$ & $file$ & $35/54$ & $19/54$ & $0$ & $0$ \\
    \end{tabular}
    \caption{Empirical model associated with the probabilistic model of $\{$\emph{coach}, \emph{boxer}$\}\times\{$\emph{lap}, \emph{file}$\}$ \label{fig:prob_coachlapboxerfile}}
\end{figure}
Recall that the model in section \ref{subsec:poss_coachlapboxerfile} was possibilistically signalling. The frequencies mined from corpora were found to have the same support as the model described in section \ref{subsec:poss_coachlapboxerfile} (see Fig. \ref{fig:prob_coachlapboxerfile}), whence the probabilistic analogue remains signalling.

\subsection{$\{$\emph{tap}, \emph{box}$\}\times\{$\emph{pitcher}, \emph{cabinet}$\}$}\label{subsec:prob_tappitcherboxcabinet}
\begin{figure}[ht]
    \centering
    \begin{subfigure}[b]{\linewidth}
        \small
        \centering
        \begin{tabular}{cc|cccc}
            $verb$ & $object$ & (0,0) & (0,1) & (1,0) & (1,1)\\\hline
            $tap$ & $pitcher$ & $17/22$ & $15/22$ & $0$ & $0$\\
            $tap$ & $cabinet$ & $1/21$ & $3/7$ & $11/21$ & $0$\\
            $box$ & $pitcher$ & $3/4$ & $1/4$ & $0$ & $0$ \\
            $box$ & $cabinet$ & $3/7$ & $10/21$ & $2/21$ & $0$ \\
        \end{tabular}
        \caption{Empirical model}    \label{subfig:prob_tapboxpitchercabinet}
    \end{subfigure}
    \begin{subfigure}[b]{\linewidth}
        \centering
        \includegraphics[width=.6\linewidth]{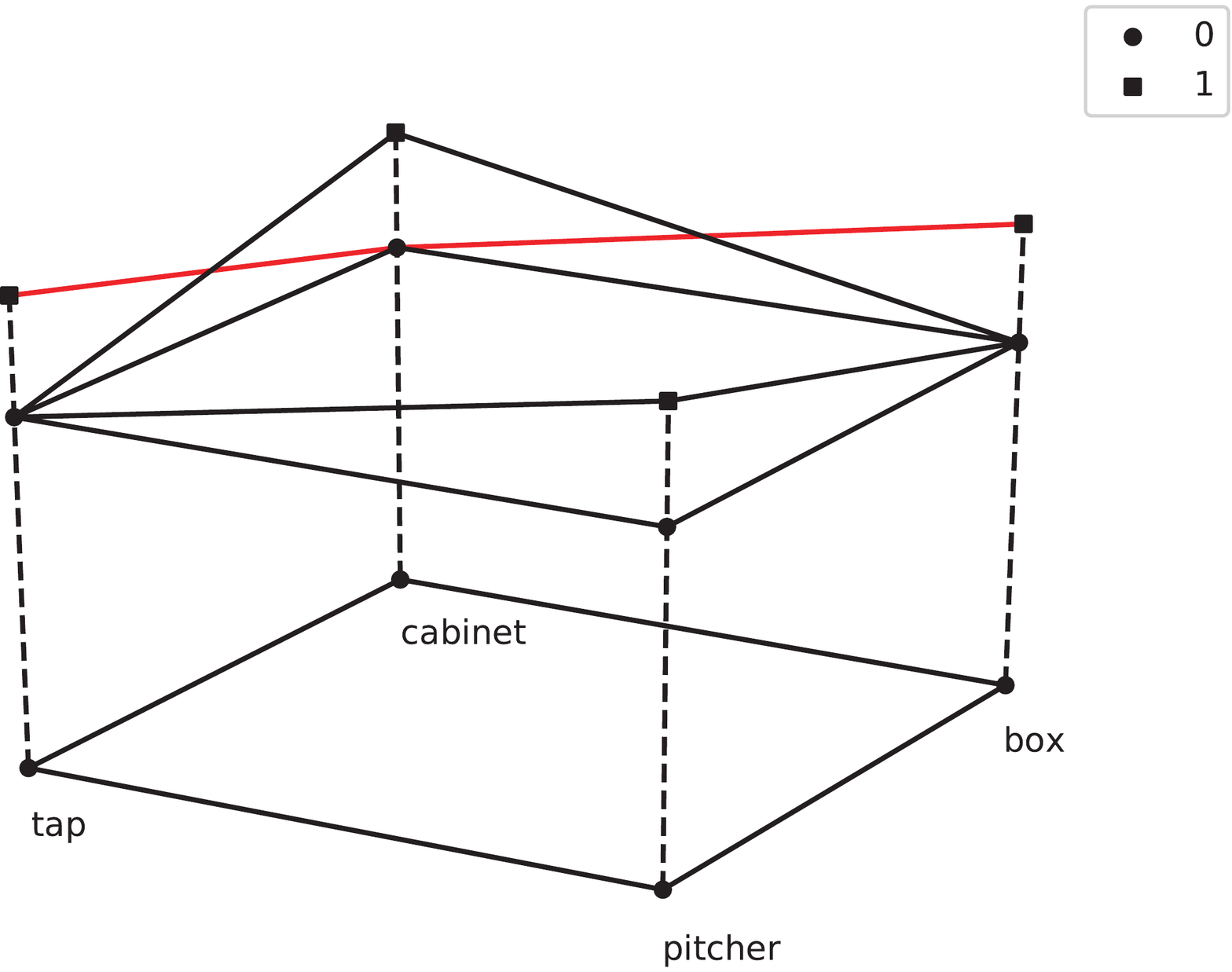}
        \caption{Bundle diagram}
    \end{subfigure}
    \caption{Probabilistic model associated with the probabilistic model of $\{$\emph{tap}, \emph{box}$\}\times\{$\emph{pitcher}, \emph{cabinet}$\}$.}
 \end{figure}

By mining frequencies of co-occurrences of phrases in our   two corpora, the model described in section \ref{subsec:poss_tappitcherboxcabinet} becomes probabilistically signalling, see  Fig.~\ref{subfig:prob_tapboxpitchercabinet}. We therefore cannot decide whether this model is probabilistically contextual in the sheaf-theoretic framework.

 It is important to note that, given the finite size and the nature of the corpora considered, many interpretations of the phrases considered did not occur; for example, there was no instance of baseball players' (pitchers')  phones or conversations being recorded (tapped). On the other hand, several other  interpretations of the phrases did occur, for example figuratively putting cabinet members in boxes or black-boxing a group of ministers. 

\subsection{$\{$\emph{press}, \emph{box}$\}\times\{$\emph{can}, \emph{leaves}$\}$}\label{subsec:prob_presscanboxleaves}

The possibilistic version of this example, presented in section \ref{subsec:poss_presscanboxleaves}, was  non-signalling. Even if tabulating the observed frequencies did not change the support of the distributions, the model has  become  probabilistically signalling. Indeed, one can check that:

\vspace{-3mm}
{\small
\begin{align}
    \label{eq:boxSignal}
         &P\left[\left.box~leaves\right|_{box}\mapsto put~in~boxes\right] &=& 2/3\nonumber \\ 
    \neq & P\left[\left.box~can\right|_{box}\mapsto put~in~boxes\right] &=& 7/74
\end{align}
}
Yet again, we cannot use the  sheaf-theoretic framework  to evaluate the contextuality of this model.
 \begin{figure}[ht]
    \centering
    \scalebox{.55}{
    \begin{tabular}{cc|ccccccccc}
        $\mathsf{A}$ & $ \mathsf{B}$ & (0,0) & (0,1) & (0,2) & (1,0) & (1,1) & (1,2) & (2,0) & (2,1) & (2,2) \\\hline
        $press$ & $can$ & $2/25$ & $0$ & $0$ & $0$ & $0$ & $41/50$ & $0$ & $1/50$ & $2/25$\\
        $press$ & $leaves$ & $0$ & $6/13$ & $0$ & $5/13$ & $0$ & $0$ & $2/13$ & $0$ & $0$\\
        $box$ & $can$ & $7/74$ & $0$ & $0$ & $0$ & $0$ & $0$ & $0$ & $1/74$ & $33/37$\\
        $box$ & $leaves$ & $0$ & $2/3$ & $0$ & $0$ & $0$ & $0$ & $1/3$ & $0$ & $0$\\
    \end{tabular}}
    \caption{Empirical model associated with the probabilistic model of $\{$\emph{press}, \emph{box}$\}\times\{$\emph{can}, \emph{leaves}$\}$.}
 \end{figure}
 
\subsection{Subject-verb v. Verb-object}\label{subsec:prob_rank2}

We now present the probability distribution arising from the examples in section \ref{subsec:poss_rank2}. As some of the previous models, the two corpora did not have instances of all the possible readings of each of the contexts. The obtained probability distributions are shown in Fig.~\ref{fig:prob_rank2}. 
As expected, the probability distribution is also signalling.

\begin{figure}[ht]
    \centering
    \begin{subfigure}[b]{\linewidth}
        \small
        \centering
        \begin{tabular}{c|cccc}
             (\emph{adopt}, \emph{boxer}) & (0,0) & (0,1) & (1,0) & (1,1) \\ \hline
             \emph{adopt} $\rightarrow$ \emph{boxer} & 0 & $29/30$ & $1/30$ & 0\\
             \emph{adopt} $\leftarrow$ \emph{boxer} & $1/4$ & 0 & 0 & $3/4$
        \end{tabular}
        \caption{\emph{adopt boxer/boxer adopts}}\label{subfig:prob_adoptboxer}
    \end{subfigure}
    \begin{subfigure}[b]{\linewidth}
        \small
        \centering
        \begin{tabular}{c|cccc}
             (\emph{throw}, \emph{pitcher}) & (0,0) & (0,1) & (1,0) & (1,1) \\ \hline
             \emph{throw} $\rightarrow$ \emph{pitcher} & $2/5$ & $0$ & $1/10$ & $1/2$\\
             \emph{throw} $\leftarrow$ \emph{pitcher} & 0 & $2/3$ & $1/3$ & 0
        \end{tabular}
        \caption{\emph{throw pitcher/pitcher throws}}\label{subfig:prob_throwpitcher}
    \end{subfigure}
    \caption{Empirical models for the pairs of words.\label{fig:prob_rank2}}
\end{figure}

\section{Non-signalling and ambiguity in natural language}\label{sec:ns}
Non-signalling is a necessary condition for demonstrating non-locality in quantum mechanics.  In experiments such as the one described in \citet{EPR}, this assumption models the space-like separation between the systems, e.g. two entangled qubits which are measured in geographically different labs, or more generally the fact that no communication between these systems is possible after their preparation. Non-signalling is a property that ensures some laws of quantum mechanics hold in specific systems and certainly there is no reason to assume it for natural language. In order to understand why not,  let's try and  use an analogy with quantum systems. In our experiment, ambiguous phrases become analogous to entangled quantum systems and  each word within a phrase to a qubit. In the \emph{subject-verb} phrases we considered, a form of communication between  words  within a phrase becomes possible if after, say the subject-measuring agent determines the meaning of the subject, the verb-measuring agent has a more limited choice in determining the meaning of the verb. A similar situation is true for the \emph{verb-object} phrases. In these cases, communication between the words of a phrase may seem possible but will definitely not  in general. For instance, consider the \emph{coach lap} phrase, if the subject-measuring agent decides that the meaning of \emph{coach} is \emph{bus}, the verb-measuring agent does not get a choice, since \emph{buses} cannot \emph{drink}.  In this case, communication between the subject and verb-measuring agents is needed.  If the subject-agent, however, sets the meaning of \emph{coach} to be  \emph{sports trainer}, the verb-measuring agent still gets a choice for the meaning of \emph{lap}, since a \emph{trainer} can \emph{run in circles} as well as  \emph{drink something up}. In this case, communication between the agents is not as clearly possible as before.

\section{Contextuality-by-Default}\label{sec:background-CbD}
We will now study the contextuality of the probabilistic signalling systems
we obtained in section \ref{sec:prob} 
using the Contextuality-by-Default framework.
In this framework, each set of jointly distributed measurements
of the empirical model is called a context, and the contextuality of
a system is defined by the impossibility of creating a global joint
distribution in which
the variables corresponding to each measurement
in each pair of contexts where they appear are equal to each other
with maximal probability (instead of always).
For example, in expression (\ref{eq:boxSignal}) we noticed
that the proportions with which the word ``box''
is assigned the meaning ``put in boxes'' differs between the contexts
with measurements ``box leaves'' and ``box can''.
This difference makes the system signalling and implies that the two
variables cannot be treated as equal to each other within a global assignment.
They need to be treated as different random variables.
The maximal probability that those two random variables can both receive the assignment
``put in boxes'' is \(\min\{2/3, 7/74\} = 7/74\).
These probabilities can be found for every pair of variables corresponding to each measurement.
Continuing with the example of the variables corresponding to the measure of ``box''
from the example in Section~\ref{subsec:prob_presscanboxleaves},
the maximal probability with which they could be assigned the meaning ``fight''
in the contexts ``box leaves'' and ``box can''
is equal to \(\min\{0, 0\} = 0\),
and the probability with which they both can be assigned ``container''
is \(\min\{1/3, 67/74\} = 1/3\).

The task of finding whether a global joint distribution
that maximizes these probabilities for every measurement exists
can be solved by linear programming.
\citet{CbD_ConteNtConteXt} describe how 
to define this task for systems that include measurements
with a finite number of outcomes by
taking all possible dichotomizations of their respective outcome sets.
We illustrate the procedure with the proportions 
of the system in Section~\ref{subsec:prob_presscanboxleaves}.
The description of this system simplifies by noting that 
the word ``leaves'' could only be assigned two meanings and that for the word ``box''

\vspace{-3mm}
{\small
    \begin{align*}
    \label{eq:dichVars}
    P\left[\left.box~leaves\right|_{box}\mapsto fight\right] &=& 0, \\ 
    P\left[\left.box~can\right|_{box}\mapsto fight\right]   &=& 0,
\end{align*}
}
effectively making those variables also binary.
Thus, we need only consider dichotomizations of variables corresponding to the measurements
for ``press'' and ``box''.

A global joint distribution of all dichotomized variables in our system
must define probabilities
for \(2^{16}\) different events.
They are the combination of the outcomes of \(16\) binary random variables:
a) \(6\) in context ``press can'' including the three dichotomizations of 
\(\left.press~can\right|_{press}\) and \(\left.press~can\right|_{can}\);
b) \(4\) in context ``press leaves'';
c) \(4\) in context ``box can''; and
d) \(2\) in context ``box leaves''.
The \(2^{16}\) probabilities are restricted by the probabilites estimated in Section~\ref{subsec:prob_presscanboxleaves}, which total
97 linear constrains considering the joint events of the dichotomizations, 
individual margins (as the ones in expression~(\ref{eq:boxSignal})), and that probabilities in a distribution add to unity.
These probabilities are further restricted by the maximal probabilities computed for the pairs of variables
corresponding to the same dichotomization of the same measurement.
These maximal probabilities are computed by taking the minimum of the two compared probabilities
as explained above, and
they amount to \(8\) linear constrains. In all, a total of \(105\) linear constrains that
the probabilities of the \(2^{16}\) events must satisfy, and that can be represented
in a \(105 \times 2^{16}\) matrix of coefficients.
Solving the set of linear equations for this example showed that it
was possible to find such a global joint distribution.
Whence, the system is not contextual.

The systems in sections \ref{subsec:prob_coachlapboxerfile} and \ref{subsec:prob_tappitcherboxcabinet} can be shown to be non-contextual within the CbD framework from a Bell inequality for certain signalling systems which was proved in \citet{CbD_conjecture}.

\subsection{Subject-verb v. Verb-object}\label{subsec:CbDcontextual}

Let us now return to the pairs of words introduced in 
\ref{subsec:poss_rank2}.
The probability distributions for the models \emph{adopt boxer/boxer adopts} and \emph{throw pitcher/pitcher throws}, mined as in section \ref{sec:prob}, are depicted in Figs.~\ref{subfig:prob_adoptboxer} and \ref{subfig:prob_throwpitcher}, respectively.

Unlike the systems considered above, these two are contextual within
the CbD framework.
This can be shown using the Bell-type inequality proved in 
\citet{CbD_conjecture} and,
using the results from \citet{Dzhafarov2020}, we can measure the degree
of contextuality of each of these two systems.
The contextuality measure for the \emph{adopt-boxer} pair is \(1/30\) and
the measure for the \emph{throw-pitcher} pair is \(7/30\).
These measures indicate how far from becoming non-contextual is each system.

Clearly, the system for the pair \emph{adopt-boxer} could easily become non-contextual
if the corpora search in the verb-object context had failed to find 
a figurative meaning of adopt, together with the fighter meaning of boxer for any
occurrence of the words ``adopt boxer''.
More generally, we can assess how reliably contextual is this system by means of parametric 
bootstrap. We find that the probability with which we could find a non-contextual system
based on the distributions in Fig.~\ref{subfig:prob_adoptboxer}
is larger than \(.56\).

The contextuality for the pair \emph{throw-pitcher} is much larger, and
indeed the system would need to exhibit many occurrences of meaning assignments
that contravene the general patterns exhibited within each of the contexts.
For example, the system would be deemed non-contextual if the proportion of times
where \emph{throw} took the literal meaning together with an interpretation of
pitcher as a \emph{jug} in the expression ``throw pitcher''
increased from \(1/10\) to \(1/3\) while preserving the overall proportions
with which each of the words was interpreted with a given meaning (say,
\emph{throw} remains interpreted literally \(3/5\) of the times).
Analogously to the previous computation, 
given the probabilities estimated in Fig.~\ref{subfig:prob_throwpitcher}
, the probability of finding the system non-contextual is larger than \(.08\).

\section{Conclusions and Discussion}
Undoubtedly, the context of ambiguous words plays an important role in their disambiguation process. The nature of this role, on the other hand, is not properly understood and quantified. In this work, we find ambiguous phrases  that are possibilistically  (i.e. logically) contextual in the sheaf-theoretic model, but show that their probabilistic extensions  become signalling. In the presence of  signalling, we  analyse these examples in the  CbD framework and discover some of them are not CbD-contextual. At the same time,  however,  we do find examples  that are CbD-contextual albeit signalling. 
 We then argue that the use of different  contextuality frameworks allows us to formally study the effect of the context on choices of interpretation of ambiguous phrases,  paving the way for a systematic study of general contextual influences in natural language.


This study was restricted by the nature of the types of meanings we considered and the size of our corpora. Indeed, the observed frequencies of phrases were not always consistent with our intuition, and in some cases, meaningful phrases did not appear in the corpus altogether. An example was the word \emph{coach}, which  could either mean \emph{a sports trainer} or a \emph{type of bus}. In the corpora we considered, the latter meaning was in fact  quite  rare. Our conjecture is that this   is due to the fact that the corpora we considered were both almost exclusively based on British English, whereas, the \emph{bus} meaning of \emph{coach} is mainly American.  Regarding types of meaning, in order to facilitate  our manual  search for occurrences of interpretations, we restricted  the domain of possible meanings  and did not consider figuratively  metaphorical options. An example is the verb \emph{boxing}, which can also mean \emph{labelling} or \emph{ignoring the workings of}, but we only considered its  \emph{putting in a box} and \emph{fist fighting} meanings. In future work, we aim to overcome this restrictions by  widening our experimental data and gather human judgement on degrees of likelihood of each meaning combination.  This will allow also us to consider a wider range of grammatical  relations in the contexts and also study the effects  of these  structures on the disambiguation process
as well as allowing a more reliable estimation of the probability distributions.

 \section*{Acknowledgments}

 This work was supported by the EPSRC Centre for Doctoral Training in Delivering quantum Technologies, grant ref. EP/S021582/1. The authors would   like to thank Ruth Kempson, Shane Mansfield, and Ehtibar Dzhafarov for  fruitful discussions. 

\bibliographystyle{acl_natbib}
\bibliography{Ref}


\end{document}